\documentclass[runningheads]{llncs}
\usepackage{amsfonts}
\usepackage{amsmath}
\usepackage{comment}
\usepackage{graphicx}
\usepackage{multirow}
\usepackage{enumerate}
\usepackage{paralist}
\usepackage{multicol}
\usepackage{cite}
\usepackage[font=small,labelfont=bf]{caption}

\begin{document}
\title{AspeRa: Aspect-based Rating Prediction Model}
%


\author{Sergey~I.~Nikolenko\inst{1,4} \and
Elena~Tutubalina\inst{1,2,4} \and
Valentin~Malykh\inst{1,3} \and
Ilya~Shenbin\inst{1}\and
Anton~Alekseev\inst{1}
}
\authorrunning{S. Nikolenko et al.}
%
\institute{Samsung-PDMI Joint AI Center, Steklov Mathematical Institute at St.~Petersburg  
\and Chemoinformatics and Molecular Modeling Laboratory, Kazan Federal University 
\and Neural Systems and Deep Learning Laboratory, \\Moscow Institute of Physics and Technology 
\and Neuromation OU, Tallinn, 10111 Estonia}



%
\maketitle              
\begin{abstract}
We propose a novel end-to-end Aspect-based Rating Prediction model (AspeRa) that estimates user rating based on review texts for the items and at the same time discovers coherent aspects of reviews that can be used to explain predictions or profile users. The AspeRa model uses max-margin losses for joint item and user embedding learning and a dual-headed architecture; it significantly outperforms recently proposed state-of-the-art models such as DeepCoNN, HFT, NARRE, and TransRev on two real world data sets of user reviews. With qualitative examination of the aspects and quantitative evaluation of rating prediction models based on these aspects, we show how aspect embeddings can be used in a recommender system.

\end{abstract}

\section{Introduction}
As the scale of online services and the Web itself grows,
recommender systems increasingly attempt to utilize texts available online, 
either as items for recommendation or as their descriptions~\cite{NA16,zheng2017joint,mitcheltree2018using,alekseev2017word}.
One key complication 
is that a single text can touch upon many different features of the item; e.g., the same brief review of a laptop can assess its weight, performance, keyboard, and so on, with different results. Hence, 
real-world applications need to separate different aspects of reviews.
This idea also has a long history~\cite{liu2012sentiment,pang2008opinion}.
Many recent works in recommender systems have applied deep learning methods~\cite{hsieh2017collaborative,seo2017representation,zheng2017joint,srivastava2017autoencoding}.
%
In this work, we 
introduce novel deep learning methods for making recommendations with full-text items, aiming to learn interpretable user representations that reflect user preferences and at the same time help predict ratings.
We propose a novel Aspect-based Rating Prediction Model (AspeRa) for aspect-based representation learning for items by encoding word-occurrence statistics into word embeddings and applying dimensionality reduction to extract the most important aspects that are used for the user-item rating estimation. We investigate how and in what settings such neural autoencoders can be applied to content-based recommendations for text items.

\section{AspeRa Model}
\begin{figure*}[t]
      \includegraphics[width=1.0\linewidth]{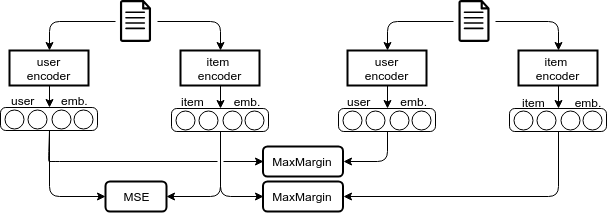}
      \vspace{-.2cm}
    \captionof{figure}{Architecture of the proposed \emph{AspeRa} model.}
      \label{fig:fig1}
  \vspace{-.3cm}
 \end{figure*}

The \emph{AspeRa} model combines the advantages of deep learning (end-to-end learning, spatial text representation) and topic modeling (interpretable topics) for text-based recommendation systems.
%
%
Fig.~\ref{fig:fig1} shows the overall architecture of AspeRa. 
The model receives as input two reviews at once, treating both identically. Each review is embedded with self-attention to produce two vectors, one for author (user) features and the other for item features. These two vectors are used to predict a rating corresponding to the review. All vectors are forced to belong to the same feature space. 
The embedding is produced by the Neural Attention-Based Aspect Extraction Model (ABAE)~\cite{abae}.
%
%
%
As in topic modeling or clustering, with ABAE the designer can determine a finite number of topics/clusters/aspects, and the goal is to find out for every document to which extent it satisfies each topics/aspects.
From a bird's eye view, ABAE is an autoencoder. The main feature of ABAE is the reconstruction loss between bag-of-words embeddings used as the sentence representation and a linear combination of aspect embeddings. A sentence embedding is 
additionally weighted by \emph{self-attention}, an attention mechanism where the values are word embeddings and the key is the mean embedding of words in a sentence.

\def\z{\mathbf{z}}
\def\r{\mathbf{r}}
\def\e{\mathbf{e}}
\def\p{\mathbf{p}}
\def\y{\mathbf{y}}
\def\b{\mathbf{b}}

The first step in ABAE is to compute the embedding $\z_s \in \mathbb{R}^d$ for a sentence $s$; below we call it a text embedding:
    $\z_s = \sum_{i=1}^n a_i \e_{w_i}$,
where $\e_{w_i}$ is a word embedding for a word $w_i$, $e \in \mathbb{R}^d$. As word vectors the authors use \emph{word2vec} embeddings trained with the skip-gram model~\cite{mikolov2013distributed}.
Attention weights $a_i$ are computed as a multiplicative self-attention model:
    $a_i = \textrm{softmax}(\e_{w_i}^\top A \y_s)$, 
where $\y_s$ is the average of word embeddings in a sentence, $\y_s = \sum_{i=1}^n \e_{w_i}$, and $A \in \mathbb{R}^{d \times d}$ is the learned attention model.
The second step is to compute the aspect-based sentence representation $\r_s \in \mathbb{R}^d$ from an aspect embeddings matrix $T \in \mathbb{R}^{k \times d}$, where $k$ is the number of aspects:
    $\p_s = \textrm{softmax}(W\z_s + \b)$,
where $\p_s \in \mathbb{R}^k$ is the vector of probability weights over $k$ aspect embeddings, $\r_s = T^\top \p_s$, and $W \in \mathbb{R}^{k \times d}$, $\b\in\mathbb{R}^k$ are the parameters of a multi-class logistic regression model. Below we call $\r_s$ the reconstructed embedding.

To train the model, ABAE uses the cosine distance between $\r_s$ and $\z_s$ with a contrastive max-margin objective function~\cite{weston2011wsabie} as the reconstruction error, also adding an orthogonality penalty term that tries to make the aspect embedding matrix $T$ to produce aspect embeddings as diverse as possible.

The proposed model's architecture includes an embedder, which provides text and reconstruction embeddings for an object similar to ABAE (``user embedding'' and ``item embedding'' on~Fig.~\ref{fig:fig1}). The intuition behind this separation of user and item embedding is as follows: there are some features (aspects) important in an item for a user, but the item also has other features. Hence, we want to extract user aspects from a user's reviews as well as item aspects from an item's reviews.
The resulting embedding is conditioned on aspect representation of the reviews; we will see below that
this model can discover interpretable topics. 
The model contains four embedders in total, one pair of user and item embedders for two reviews being considered at once, as shown on Fig.~\ref{fig:fig1}. 
First each review is paired with another review of the same user, grouping by users and shuffling the reviews inside a group; then with another review of the same item. Thus, the training set gives rise to only twice as many pairs as reviews available for training. The rating score for the first review in a pair is used to train the rating predictor (\textit{MSE}); at prediction stage, only one ``tower'' is used.

There are two losses in \emph{AspeRa}: MSE for rating prediction (Fig.~\ref{fig:fig1}) and MaxMargin loss to put user and item embeddings in the same space (Fig.~\ref{fig:fig1}).
The MSE loss assumes that rating is predicted as the dot product of user and item embeddings for a review:
    $MSE = \frac{1}{N} \sum_{j=1}^N ({\z^u_j}^\top \z^i_j - r_j)^2$, 
where $\z^u_j$ is a text embedding for the author of review $j$, $\z^i_j$ is a text embedding for the item $j$ is about, and $r_j$ is the true rating associated with $j$. 
Max-margin loss aims to project all user and item embeddings into the same feature (aspect) space; see Fig.~\ref{fig:fig1}. We use it in two ways. First, we push reconstructed and text embeddings to be closer for each user $i$, and pushes text embeddings for both considered items apart:
$    \mathrm{MaxMargin}(i, j) = \frac{1}{N} \sum_{i,j} \max(0, 1 - {\r^u_i}^\top \z^u_i +  {\r^u_i}^\top \z^i_i + {\r^u_i}^\top \z^i_j),\label{eq:mm1}
$
where $i,j$ are indices of reviews, $\r^u_i$ is a reconstructed embedding from ABAE for user $i$, $\z^u_i$ is a text embedding for user $i$,  $\z^i_i$ and $\z^i_j$ are text embeddings from ABAE for items $i$ and $j$ respectively. 
This loss is applied for all four possible combination of users and items, i.e., $(u_i,i_i,i_j), (u_j,i_i,i_j), (i_i,u_i,u_j), (i_j,u_i,u_j)$.
Second, we keep user embeddings from two reviews of the same author close:
%
$    \mathrm{MaxMargin}(i, j) = \frac{1}{N} \sum_{i,j} \max(0, 1 - {\z^u_i}^\top \z^u_j +  {\z^u_i}^\top \z^i_i + {\z^u_i}^\top \z^i_j),\label{eq:mm2}
$
where $i,j$ are indices of reviews, $\z^u_i$ and $\z^u_j$ are user embeddings from ABAE for authors of reviews $i$ and $j$ and $\z^i_i$ and $\z^i_j$ are text embeddings from ABAE for items $i$ and $j$ respectively. This second form is symmetrically applied to item and user embeddings for two reviews pf the same item from different authors.

\section{Experimental Evaluation} \label{sec:exp}

\subsubsection{Datasets and experimental setup.} 
We evaluated the proposed model on \emph{Amazon Instant Videos 5-core reviews} and \emph{Amazon Toys and Games 5-core reviews}\footnote{http://jmcauley.ucsd.edu/data/amazon/}~\cite{he2016ups,mcauley2015image}. The first dataset consists of reviews written by users with at least five reviews on \emph{Amazon} and/or for items with at least five reviews; it contains $37{,}126$ reviews, $5{,}130$ users, $1{,}685$ items, and a total of $3{,}454{,}453$ non-unique tokens. 
The second dataset  follows 5 minimum reviews rule; it contains $167{,}597$ reviews,  $19,412$ users, $11,924$ items, and a total of $17,082,324$ non-unique tokens.
We randomly split each dataset into $10$\% test set and $90$\% training set, with $10\%$ of the training set used as a validation set for tuning hyperparameters.
%
%
%
Following ABAE~\cite{abae}, we set the aspects matrix ortho-regularization coefficient equal to $0.1$. Since this model utilizes an aspect embedding matrix to approximate aspect words in the vocabulary, initialization of aspect embeddings is crucial.
The work~\cite{he2017unsupervised} used \emph{k-means} clustering-based initialization~\cite{steinhaus1956division,macqueen1967some,lloyd1982least}, where 
the aspect embedding matrix is initialized with centroids of the resulting clusters of word embeddings.
%
We compare two word embeddings for \emph{AspeRa}: \emph{GloVe}~\cite{Pennington2014} and \emph{word2vec}~\cite{MCCD13,NIPS2013_5021}. We adopted a \emph{GloVe} model trained on the \emph{Wikipedia~2014~+~Gigaword~5} dataset (6B tokens, 400K~words vocabulary, uncased~tokens) with dimension~$50$. For \emph{word2vec}, we used the training set of reviews to train a skip-gram model (SGNS) with the \emph{gensim} library~\cite{rehurek2010gensim} with dimension $200$, window size~$10$, and $5$~negative samples; see Table~\ref{tab:aspera-params} for details.

\subsubsection{Rating Prediction.}\label{ssec:rating}

\begin{figure*}[t]
\begin{minipage}[!t]{0.45\textwidth}
\captionof{table}{Two sets of AspeRa hyperparameters (for models with different initialization strategies).}
\renewcommand{\tabcolsep}{3pt}
\label{tab:aspera-params}
\centering
\begin{tabular}{|l|p{1.5cm}|p{1.5cm}|}
\hline
\textbf{Settings} & \textbf{AspeRa (GloVe)} &\textbf{AspeRa (SGNS)} \\  \hline
 Embeddings & GloVe & SGNS  \\ 
 Optimization alg. & Adam \cite{DBLP:journals/corr/KingmaB14}  & Adam \\
 \# aspects & 11  & 10 \\
 Hidden layer dim. & 256 & 64 \\
 \# epochs &  20  & 18 \\
 \# words per sample & 256 & 224 \\ 
\hline
\end{tabular}

\end{minipage}
      \hfill
\begin{minipage}[!t]{0.48\textwidth}
\captionof{table}{Performance of text-based and collaborative rating prediction models.}
\label{tab:MSE}
\renewcommand{\tabcolsep}{3pt}\scriptsize
\centering
\begin{tabular}{|l|p{1.1cm}|p{1.2cm}|}
\hline
\multirow{2}{*}{\textbf{Model}} & \multicolumn{2}{|c|}{\textbf{MSE}}  \\ 
\cline{2-3}
 & \textbf{Instant Videos} & \textbf{Toys \& Games} \\ \hline
NMF  & 0.946 & 0.821\\ 
DeepCoNN   & 0.943 & 0.851\\ 
Attn+CNN& 0.936 & -\\ 
SVD & 0.904 &  0.788\\ 
HFT  & 0.888 & 0.784\\ 
TransRev  & 0.884 & 0.784\\
NARRE & - & 0.769 \\
AspeRa (GloVe)   & 0.870 & 0.730 \\
AspeRa (SGNS)    & \textbf{0.660} & \textbf{0.571}\\ \hline
\end{tabular}
\end{minipage}\vspace{.03cm}

\includegraphics[width=\linewidth,keepaspectratio]{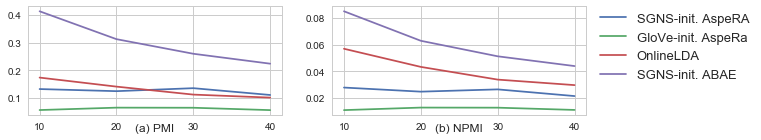}
    \caption{Comparing \emph{AspeRa} with GloVe (SGNS clusters), ABAE (SGNS clusters), and LDA with the same vocabulary and 10~topics on \emph{Instant Videos}; more is better. X-axis: number of top-ranked representative words per aspect, Y-axis: topic coherence scores.} 
    \label{fig:fig-topic-quality}
\end{figure*}

We evaluate the performance of \emph{AspeRa} in comparison to state-of-the-art models: NMF \cite{zhang2006learning}, DeepCoNN  \cite{zheng2017joint}, Attn+CNN \cite{seo2017representation}, SVD \cite{KBV09}, HFT \cite{mcauley2013hidden}, NARRE \cite{NARRRE2018}, and TransRev \cite{transrev}; we introduce these models in Section~\ref{sec:rw}. 
Table~\ref{tab:MSE} compares the best Mean Square Error (MSE) of \emph{AspeRa} and other models for rating prediction. Results of existing models were adopted from \cite{transrev} for \emph{Amazon Instant Videos 5-core reviews} with the ratio 80:10:10. We also used the results of NARRE model~\cite{NARRRE2018}, obtained in the same setup as~\cite{transrev} but with a different random seed.
Note that while \emph{AspeRa} with generic \emph{GloVe} word embeddings still works better than any other model, adding custom word embeddings trained on the same type of texts improves the results greatly.

\subsubsection{Topic Quality}\label{ssec:topics}



We compared the performance of \emph{AspeRa} with \emph{OnlineLDA}~\cite{hoffman2010online} trained with the \emph{gensim} library~\cite{rehurek2010gensim}, with the same vocabulary and number of topics, and ABAE with $10$ aspects and $18$ epochs, initialized with the same \emph{word2vec} vectors (SGNS) as \emph{AspeRa} and having the same ortho-regularization coefficient as the best \emph{AspeRa} model, evaluating the results in terms of topic coherence metrics, NPMI~\cite{bouma2009normalized} and PMI~\cite{newman2009external,Newman2010} computed with companion software for~\cite{lau2014machine}.
Figure~\ref{fig:fig-topic-quality} shows that the quality is generally lower for larger number of representative words per aspect (horizontal axis), and that \emph{AspeRa} achieves scores comparable to LDA and ABAE, although ABAE remains ahead.
Tables~\ref{tab:deepcoabae-w2v-topics} and \ref{tab:deepcoabae-glove-topics} present several sample aspects discovered by \emph{AspeRA}.
Qualitative analysis shows that some aspects describe what could be called a \textit{topic} (a set of words diverse by part of speech and function describing a certain domain), some encode sentiment (top words are adjectives showing attitude to certain objects discussed in the text), and some encode names (actors, directors, etc.). We also found similar patterns in the output of the basic ABAE model~\cite{abae}. Thus, most aspects are clearly coherent, but there is room for improvement.



\begin{table}[t]
\caption{Sample aspects from \emph{Instant Videos} discovered by AspeRa (SGNS).}
\label{tab:deepcoabae-w2v-topics}
\begin{center}
\begin{tabular}{|l|p{11.5cm}|}
\hline
\# & \textbf{Aspect words}    \\ \hline
1 & communities governments incidents poverty unity hardships slaves citizens fought \\
2 & coppola guillermo bram kurosawa toro ridley del prolific ti festivals  \\ 
3 & brisk dialouge manipulation snappy plotlines dialogues taunt camerawork muddled  \\
4 & sock vegans peanut stifling bats buh ammonium trollstench vegetables pepsi  \\ 
5 & the a and to is of joe's enters that fatal  \\ \hline
\end{tabular}
\end{center}

\caption{Sample aspects from \emph{Instant Videos} discovered by AspeRa (GloVe).}
\label{tab:deepcoabae-glove-topics}
\centering
\begin{tabular}{|l|l|}
\hline
\# & \textbf{Aspect words} \\ \hline
1 & protein diagnose cell genes brain membrane interacts interact oxygen spinal \\
2 & boost monetary raise introduce measures credit expects increase push demand  \\
3 & towel soaked greasy towels cloth dripping tucked crisp coat buckets  \\
4 & offbeat comic parody spoof comedic quirky cinematic campy parodies animated  \\
5 & sheesh wham whew hurrah oops yikes c'mon shhh oooh och \\ \hline
\end{tabular}
\end{table}

\section{Related Work} \label{sec:rw}

Classical collaborative filtering based on matrix factorization (MF)~\cite{KBV09,zhang2006learning} has been extended with textual information, often in the form of topics/aspects;
aspect extraction uses topic modelling~\cite{Zhu06,tutubalina2015inferring,tutubalina2016constructing} and phrase-based extraction~\cite{solovyev2014dictionary}.
Collaborative topic regression (CTR)~\cite{wang2011collaborative} was one of the first models to combine collaborative-based and topic-based approaches to recommendation;
to recommend research articles; it uses an LDA topic vector as a prior of item embeddings for MF.
Hidden Factors and Hidden Topics (HTF)~\cite{mcauley2013hidden} 
also combines MF and LDA but with user reviews used as contextual information.
A few subsequent works use MF along with deep learning approaches; e.g., Collaborative Deep Learning (CDL)~\cite{wang2015collaborative} improves upon CTR by replacing LDA with a stacked denoising autoencoder.
Unlike our approach, all these models learn in alternating rather than end-to-end manner. 
Recent advances in distributed word representations have made it a cornerstone of modern natural language processing~\cite{Goldberg15c}, with
neural networks recently used to learn text representations. 
He~et~al.~\cite{abae} proposed an unsupervised neural attention-based aspect extraction (ABAE) approach that
encodes word-occurrence statistics into word embeddings and applies an attention mechanism to remove irrelevant words, learning a set of aspect embeddings. 
Several recent works, including DeepCoNN~\cite{zheng2017joint}, propose a completely different approach. DeepCoNN is an end-to-end model, both user and item embedding vectors in this model are trainable functions (convolutional neural networks) of reviews associated with a user or item respectively. Experiments on \emph{Yelp} and \emph{Amazon} datasets showed significant improvements over HFT.
\emph{TransNet}~\cite{catherine2017transnets} adds a regularizer on the penultimate layer that forces the network to predict review embedding. \emph{TransRev}~\cite{transrev} is based on the same idea of restoring the review embedding from user and item embeddings. \emph{Attn+CNN} and \emph{D-Attn}~\cite{seo2017representation,seo2017interpretable} extend \emph{DeepCoNN} with an attention mechanism on top of text reviews; it both improves performance and allows to explain predictions by highlighting significant words. However, user and item embeddings of these models are learned in a fully supervised way, unlike the proposed model. Our model combines semi-supervised embedding learning, which makes predictions interpretable similar to HTF, with a deep architecture and end-to-end training.

\section{Conclusion}
We have introduced a novel approach to learning rating-~and text-aware recommender systems based on ABAE, metric learning, and autoencoder-enriched learning.  Our approach jointly learns interpretable user and item representations. It is expectedly harder to tune to achieve better quality, but the final model performs better at rating prediction and almost on par at aspects coherence with other state-of-the-art approaches. Our results can also be viewed as part of the research effort to analyze and interpret deep neural networks, a very important recent  trend~\cite{kadar2017representation,radford2017learning}.
%
%
We foresee the following directions for future work: 
\begin{inparaenum}[(i)]
\item  further improving prediction quality (especially for models that learn interpretable user representations),
\item integrating methods that can remove ``purely sentimental'' aspects into interpretable models for recommendations that we have discussed above,
\item developing visualization techniques for user profiles.
\end{inparaenum}

\paragraph*{Acknowledgements.}
This research was done at the Samsung-PDMI Joint AI Center at PDMI RAS and was supported by Samsung Research.

%
%

\end{document}